\documentclass{article}

\PassOptionsToPackage{numbers, compress}{natbib}

    \usepackage[preprint]{neurips_2024}

\usepackage[utf8]{inputenc} 
\usepackage[T1]{fontenc}    
\usepackage{hyperref}       
\usepackage{url}            
\usepackage{booktabs}       
\usepackage{amsfonts}       
\usepackage{nicefrac}       
\usepackage{microtype}      
\usepackage{xcolor}         
\usepackage{amsmath}
\usepackage{amssymb}
\usepackage{pifont}
\usepackage{kotex}
\usepackage{graphicx}

\usepackage{multirow}
\usepackage{algorithm}
\usepackage{algpseudocode}

\newcommand\highlight[1][yellow]{%
  \bgroup
  \markoverwith{\textcolor{#1}{\vrule width.1em height.8em depth.2em}}%
  \ULon
}

\title{
Toward Efficient Permutation for Hierarchical N:M Sparsity on GPUs}

\author{%
  Seungmin Yu \\
  Sungkyunkwan University\\
  \texttt{sminyu@skku.edu} \\
  \And
  Xiaodie Yi\\
  Sungkyunkwan University \\
  \texttt{cindy.135luv@skku.edu}
  \And
  Hayun Lee \\
  Sungkyunkwan University \\
  \texttt{lhy920806@skku.edu}
  \And
  Dongkun Shin\thanks{Corresponding author.} \\
  Sungkyunkwan University\\
  \texttt{dongkun@skku.edu} \\
}

\begin{document}
\maketitle

\begin{abstract}
N:M sparsity pruning is a powerful technique for compressing deep neural networks (DNNs), utilizing NVIDIA's Sparse Tensor Core technology. This method benefits from hardware support for sparse indexing, enabling the adoption of fine-grained sparsity to maintain model accuracy while minimizing the overhead typically associated with irregular data access. 
Although restricted to a fixed level of sparsity due to its reliance on hardware, N:M sparsity can be combined with coarser sparsity techniques, such as vector-wise sparsity, to achieve diverse compression ratios. 
Initially, column-wise vector sparsity is applied to a dense model, followed by row-wise N:M sparsity on the preserved column vectors. We call this multi-level approach as hierarchical N:M (HiNM) sparsity. 
Similar to earlier single-level sparsity techniques, HiNM sparsity necessitates an effective channel permutation strategy to maximize the accuracy of the compressed networks. 
However, it introduces further complexities by requiring the rearrangement of both input and output channels, addressing challenges such as permutation sequence, HiNM-sparsity-aware permutation, and maintaining consistency in channel ordering across layers. 
In this paper, we introduce a channel permutation method designed specifically for HiNM sparsity, named gyro-permutation. 
This method is crafted to exploit the unique characteristics of HiNM pruning, incorporating a strategic policy in each permutation phase, including channel sampling, clustering, and assignment, to circumvent local minima. 
Additionally, we have developed a GPU kernel that facilitates independent layer permutation during the execution of HiNM sparse networks. 
Our extensive experimental evaluations on various DNN models demonstrate that our gyro-permutation significantly enhances the accuracy of HiNM sparse networks, allowing them to reach performance levels comparable to those of unstructured sparse networks.
\end{abstract}

\section{Introduction}
Deep Neural Networks (DNNs) have been rapidly increasing in size, supporting the Scaling Law~\cite{scaling_law} that suggests larger networks usually yield higher accuracy. For instance, an image task model, DALL-E, contains approximately 12 billion parameters, and GPT-3 has over 175 billion parameters. However, this growth significantly raises the costs associated with memory, storage, and computation, creating substantial challenges for deploying DNNs on standard hardware for practical applications.

Weight pruning, which zeros out less critical network elements, has emerged as a vital solution to these challenges. Among various pruning methods~\cite{han_unstructured, channel_pruning, ovw, filter_pruning}, N:M pruning~\cite{STC}, which zeros out all but N elements of every M consecutive weight elements, has gained attention due to its hardware-supporting sparse indexing and fine granularity of the sparsity pattern. NVIDIA's Sparse Tensor Core (STC) supports the execution of N:M sparse networks, without requiring additional software-level indexing overhead. N:M sparsity, similar to element-wise unstructured sparsity, can maintain high accuracy by individually zeroing out weight elements within a given range.

\begin{figure}[htb]
  \centering
  \includegraphics[width=1\textwidth]{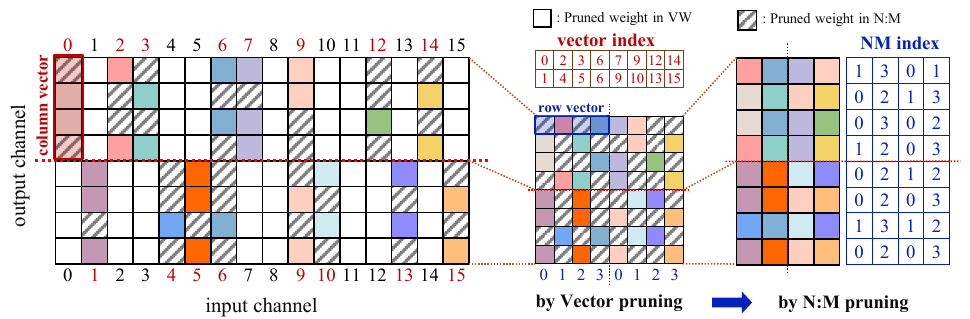}
  \caption{Hierarchical N:M sparsity pattern}
  \label{Hierarchical N:M sparsity pattern}
\end{figure}

Although the current hardware supports only a fixed level of N:M sparsity (e.g., 2:4 sparsity), various compression ratios can be achieved by combining N:M sparsity with a coarser-level sparsity, such as channel-wise or vector-wise sparsity. 
For instance, a $V \times 1$ column-wise vector sparsity is first applied, followed by row-wise N:M sparsity,
which we call this combined sparsity pattern \textbf{hierarchical N:M (HiNM) sparsity}. 
Figure~\ref{Hierarchical N:M sparsity pattern} illustrates the compression process for HiNM sparsity patterns. 
The dense weight matrix is first compressed with 50\% sparsity through 4$\times$1 column-wise vector pruning, and the corresponding vector index is generated. 
Subsequently, 2:4 pruning is applied, where two elements from each 1$\times$4 row vector are eliminated, generating the NM index to represent the location of the selected weight elements within the corresponding row vector. 
Then, the final sparsity becomes 75\% ($1 - (1 - 0.5) \times 0.5$). 
While the vector index is utilized by GPU kernel code to load only the corresponding input data from global memory into shared memory, the NM index is used by the hardware to load the corresponding input data from shared memory into the computational units.  

Since $V \times 1$ vector pruning and N:M pruning impose constraints on the sparsity pattern, channel permutation can alleviate the accuracy drop caused by these pruning techniques. 
For the column-wise vector pruning, we can rearrange the order of output channels by clustering elements with similar importance scores into a same vector~\cite{shfl-bw,ovw}.
For the row-wise N:M pruning, input channel rearrangement is required to achieve a balanced distribution of important elements among row vectors~\cite{nvidia_channelperm}.

HiNM pruning necessitates an effective permutation algorithm. 
However, it differs from permutations used for single-level sparsity pruning in several key aspects: 
\begin{enumerate}
    \item Hierarchical Pruning Awareness: Since prunings are applied to both output and input dimensions in HiNM technique, an optimal channel permutation at one dimension may not be the best solution overall. For instance, while an output channel permutation may consolidate important elements into the same vector, subsequent N:M pruning might eliminate some of these preserved elements. Consequently, channel permutation for HiNM presents significant challenges.
    \item Consistency across Layers: HiNM pruning involves permutations in both input and output channel dimensions, requiring consistency in channel orders between adjacent layers. More specifically, the output channel order of the \(i\)-th layer must align with the input channel order of the \((i+1)\)-th layer.
    \item Escaping Local Minima: We observed that existing permutation techniques can easily fall into local minima. Addressing this challenge is essential for enhancing the performance of HiNM pruning.
\end{enumerate}

In order to address the outlined issues, it is necessary to enhance the channel permutation process for HiNM pruning, achieving higher network accuracy. 
In this paper, we introduce a novel permutation technique for HiNM sparsity, termed \textbf{Gyro-Permutation}, which involves permutations for both output and input channels. 
Gyro-permutation is composed of distinct steps, including channel sampling, clustering, and assignment, as each tailored to address the challenges of hierarchical pruning awareness and avoidance of local minima.
Additionally, we propose the GPU kernel for HiNM sparse networks that rearranges input channel order at runtime. It enables two sequential layers to have different channel orders, effectively eliminating the issue of maintaining channel order consistency across layers. 

Experimental results on various DNN models confirm that 
gyro-permutation has the ability to improve the accuracy of networks significantly when employing HiNM sparsity. 
Furthermore, our technique enables HiNM sparse networks to attain accuracy levels comparable to those of unstructured sparse networks, illustrating its effectiveness and potential for widespread application in deep learning optimizations.

\section{Related Work}

\textbf{Weight pruning} Given the vast number of parameters in modern DNN models, weight parameter pruning is crucial to reach the constraints on computation and memory resources.
Various weight pruning strategies have been developed, incorporating different sparsity patterns~\cite{channel_pruning,deepcomp,filter_pruning}, salience score estimation methods~\cite{movement,mfac,obert}, and fine-tuning techniques such as one-shot pruning~\cite{oneshot} and gradual pruning~\cite{obert}.

Typically, irregular and fine-grained sparsity patterns minimize the accuracy loss associated with pruning but increase the overhead of sparse indexing. 
Vector-wise sparsity~\cite{shfl-bw,ovw,balanced_vw}, which prunes vectors of shape \(V\times1\), offers an effective balance between accuracy drop and indexing overhead. 

In case of N:M sparsity, \(N\) weight elements in a \(1\times M\) vector can be selected irregularly. However, thanks to the hardware-level indexing capabilities of NVIDIA's Sparse Tensor Core (STC), the overhead of sparse indexing can be substantially reduced.
This kind of advantage has spurred numerous research initiatives on N:M sparsity, including the development of effective training methods~\cite{learning_nm, transposable_nm}, layer-wise sparsity~\cite{domino}, and optimized GPU kernel designs~\cite{dissecting}.

Currently, since STC supports a fixed sparsity ratio, some researchers proposed new hardware designs for hierarchical sparsity~\cite{highlight,tstc} to enable various levels of sparsity. 
These designs accommodate N:M sparsity in a layered approach but often involve complex and costly hardware modifications. 
We can also consider a software-based hierarchical sparsity technique. 
Venom~\cite{venom} combines vector-wise sparsity and N:M sparsity, 
while the sparse index for vector-wise sparsity is handled by software, and N:M sparsity is managed by hardware. 
We target the software-based hierarchical sparsity and propose a channel permutation technique to minimize the accuracy drop. 

\textbf{Channel permutations} Channel permutation techniques~\cite{shfl-bw,ovw,nvidia_channelperm,tetris} preprocess weight matrices by reordering output or input channels. This reorganization aligns with the required sparsity pattern, enabling more efficient removal of unimportant elements. It involves partitioning the channels according to the target sparsity pattern and grouping appropriate channels within each partition to facilitate the pruning process. 

In previous research, specific permutation strategies have been developed for single-level sparsity patterns such as column-wise vector sparsity and N:M sparsity. For column-wise vector sparsity, output channel permutation~\cite{ovw} employs a balanced K-means clustering algorithm to categorize channels with similar distributions, thereby concentrating less significant elements for vector-by-vector removal. Conversely, input channel permutation~\cite{nvidia_channelperm} for N:M sparsity utilizes a channel swapping technique to balance the distribution of significant elements across each row vector.

Additionally, Tetris~\cite{tetris} tackles the challenge of rearranging both the output and input channels of weight matrices for block-wise sparsity. Similar to the approach in ~\cite{nvidia_channelperm}, Tetris implements channel swapping to modify the order of channels along different axes. However, Tetris introduces further index translation operations between layers to manage inconsistent channel orders, which significantly increases the overhead during GPU inference.

Our gyro-permutation differentiates itself by seamlessly integrating index translation operations into the native indexing process of the HiNM sparsity pattern. This integration occurs during the transfer of column-wise vectors from global to shared GPU memory, effectively eliminating any additional computational overhead. This novel approach not only simplifies the computational process but also enhances the overall efficiency and scalability of sparsity implementations in neural networks, as demonstrated in our experiments.

\section{Hierarchical N:M sparsity}
\subsection{Pruning policy of multi-level sparsity} \label{subsec:Pruning policy}
Hierarchical N:M(HiNM) sparsity is established through a multi-layered pruning approach, comprising column-wise vector pruning followed by N:M pruning, allowing the adoption of varied pruning strategies. 
Practitioners can opt to initiate with column-wise vector pruning, commence with N:M pruning, or integrate both methods concurrently. 
The sequence from column-wise vector pruning to N:M pruning, similar to the VENOM~\cite{venom} method, is our standard approach,
as it directly facilitates the generation of hierarchical N:M sparsity patterns. 
Conversely, beginning with N:M pruning requires subsequent adjustments for column-wise vector pruning, an unconventional strategy given the initial focus on finer granularity. 
Simultaneously considering both pruning methods necessitates a novel strategy that addresses the structural inter-dependencies of each approach. We adopt a foundational pruning policy that progresses from column-wise vector pruning to N:M pruning, demonstrating that our gyro-permutation technique achieve high performance without considering the pruning policy.

\subsection{Sparse matrix multiplication with HiNM sparsity patterns}

\begin{figure}[htb]
  \centering
  \includegraphics[width=1\textwidth]{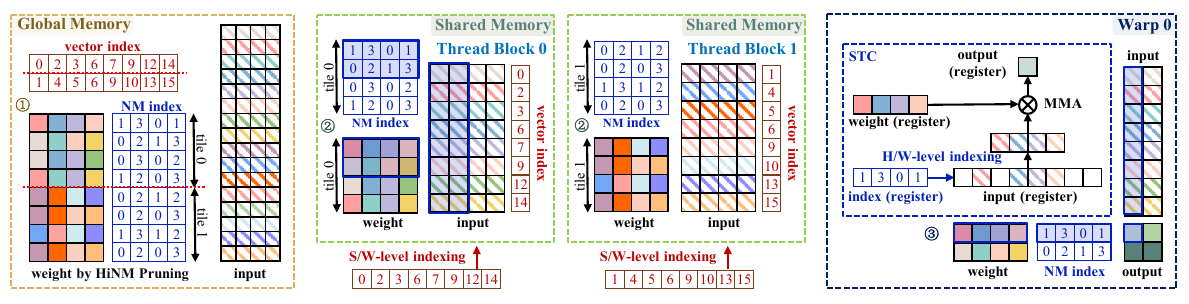}
  \caption{SpMM operation of HiNM sparsity}
  \label{SpMM operation of HiNM sparsity}
\end{figure}

In the GPU-based sparse matrix multiplication (SpMM) process involving a sparse weight matrix with HiNM sparsity patterns and a dense input matrix, a pivotal observation is the movement of column vector data. 
As shown in Figure~\ref{SpMM operation of HiNM sparsity}, during the transition from global memory to shared memory, input data is loaded according to the indices of the corresponding column vectors which are aligned along the input channel axis-from \ding{172} to \ding{173}. 
This alignment enhances memory consistency and maximizes input reuse during computation, both essential for efficient model inference.

Moreover, to fully exploit the parallel computing capabilities of GPUs, each thread block processes a "tile"—a collection of contiguous output channels equal in size to the column vector. As demonstrated in Figure~\ref{SpMM operation of HiNM sparsity}, two distinct tiles are assigned to different thread blocks. This arrangement guarantees that as each tile is computed independently, altering the order of vector indices within a tile does not impact the final computation results. This insight is advantageous for two primary reasons:

\textbf{Tile-wise input channel permutation}: By focusing permutation efforts within the granularity of column vectors in tiles rather than altering the order of larger input channels, optimization is more readily attainable. This method simplifies the permutation process and reduces the complexity associated with reaching optimal configurations.

\textbf{Maintaining layer consistency}: Traditionally, modifying the sequence of both output and input channels~\cite{tetris} necessitates additional permute operations during runtime to preserve consistency across different layers. However, if adjustments are made to the order of output channels and column vectors, these can be configured offline. By pre-ordering all layers according to the output channel sequence, during runtime, input data are loaded from global to shared memory using the reordered vector indices, thus obviating the need for additional indexing operations.

\section{Gyro-Permutation}
\label{sec:Gyro}
\subsection{Optimization challenges in permutation}
\label{subsec:Optimizing Permutations}
In a layer of a pretrained dense model, a pruning method applies a mask $M$ to a weight matrix $W \in \mathbb{R}^{m \times n}$, aiming to maximize the retention of important weight elements based on their saliency while complying with specific structural constraints.
The hierarchical N:M (HiNM) sparsity pattern, characterized by a two-step pruning process of column-wise vector pruning followed by row-wise 2:4 pruning, presents considerable challenges. This approach necessitates the permutation of both output and input channels, significantly increasing the complexity of determining optimal permutations.

Given the column-wise vector mask $M_v$ and row-wise 2:4 mask $M_{24}$, we define this challenge as the following optimization problem for applying permutations in the HiNM structure:

\begin{equation}
\underset{\sigma_o,\sigma_i^0,...,\sigma_i^T}{\operatorname{argmax}}\left\|M \odot\rho\left[\sigma_0 ; \sigma_i\right]\right\| \text { s.t. } M \text { satisfies } M_v \text { and } M \setminus \{0:0\in M_v\} \text { satisfies } M_{24}
\end{equation}

Where $\sigma_o$ represents the order of output channels, and $\sigma_i$ denotes a set of column-wise vector orders per tile $\sigma_i^0, ..., \sigma_i^T$, with $T$ being the number of tiles, and $\rho$ signifies the saliency score of the weights.
The potential number of permutations, calculated as $\frac{m!}{V!^{P_o}P_o!} \times T \times \frac{n!}{4!^{P_i}P_i!}$ where $V$ specifies the size of the column vectors and $P_o$ and $P_i$ indicate the number of partitions for the output channels and column-wise vectors, respectively, demonstrates that even for a relatively small $16 \times 16$ weight matrix with a column vector size of $4$, there could be more than 27 trillion possible permutations.

To streamline the search for effective permutations within the HiNM structure, we leverage the insight that the reordering of output channels and column-wise vectors serve distinct goals—enhancing the efficiency of column-wise vector pruning and N:M pruning, respectively. This approach is structured by sequentially transitioning from column-wise vector pruning to N:M pruning, as detailed in section~\ref{subsec:Pruning policy}. This realization allows us to decompose the overall problem into two sub-problems: output channel permutation and tile-wise input channel permutation.
Initially, the output channel permutation focuses on rearranging the output channels to ensure that the less salient elements are targeted first in the vector pruning:

\begin{equation}
    \underset{\sigma_0}{\operatorname{argmax}}\left\|M_v \odot \rho\left[\sigma_0 ; I\right]\right\| \text { s.t. } M_v \text { satisfies column-wise vector constraints }
\end{equation}

This reordering ensures that less critical weights are pruned first. Notably, this step does not require a complete segregation of important and unimportant elements into distinct column vectors, as even the elements retained after vector-wise pruning may still be targeted in subsequent N:M pruning steps.

Following the reordering of output channels, the next step optimizes the arrangement of column-wise vectors within each tile to ensure that the weight distribution aligns with the desired uniform sparsity for each row-wise vector:

\begin{equation}
    \underset{\sigma_i^0,...,\sigma_i^B}{\operatorname{argmax}}\left\|M_{24} \odot (M_v \odot \rho\left[\sigma_o ; I\right])[I; \sigma_i]\right\| \text { s.t. } M_{24} \text { satisfies 2:4 structure constraints }
\end{equation}

These permutations are conducted in a sequential manner, beginning with the output channel permutation. Subsequently, the weight matrix undergoes compression via column-wise vector pruning, followed by tile-wise input channel permutations.
This sequence aligns with our HiNM structured pruning steps, ensuring that each stage is optimally configured for effective sparsification.

\subsection{Permutation algorithm}
\label{subsec:Gyro_comp}
In the development of the Gyro-Permutation technique for HiNM sparsity, we delineated two sub-problems as outlined in section~\ref{subsec:Optimizing Permutations}: output channel permutation and tile-wise input channel permutation.
These sub-problems address distinct aspects of the sparsity structure yet are managed through a unified algorithmic framework within the Gyro-Permutation method.

Gyro-Permutation operates in three critical phases in an iteration step: sampling, clustering, and assignment, each pivotal to the permutation process.

\textbf{Sampling.} During the sampling process, we consistently extract an equal number of channels from all partitions to promote global optimization at each iteration. The effectiveness of permutations is significantly influenced by the number of samples extracted from each partition, akin to the effect of learning rates in model training. Generally, extracting a larger number of samples aids in avoiding local minima but may hinder achieving the absolute optimum. Conversely, extracting fewer samples can facilitate reaching the optimum but at an increased risk of encountering local minima. In the case of output channel permutation, we dynamically adjust the number of samples from each partition in every iteration, analogous to how learning rates are adjusted during model training. For tile-wise input channel permutation, where each partition typically contains only four column vectors, we are constrained to extract just one column vector per partition. This restriction is informed by the reduced likelihood of falling into local minima in this phase, which diminishes the need to vary the number of samples.

\textbf{Clustering.} Clustering is strategically employed to synchronize the number of sampled channels with the partition count. In tile-wise input channel permutations, where the sample count naturally aligns with the partition number, the clustering phase is bypassed, thereby simplifying the procedure. Conversely, for output channel permutation, we adopt an approach informed by prior research~\cite{ovw}. In this approach, sampled output channels are organized using the Balanced K-means clustering algorithm. This technique groups channels with similar weight distributions, enhancing the probability of aggregating less critical elements together.

\textbf{Assignment.} During this phase, samples that have been clustered are placed within the designated partitions based on a carefully defined cost function. This function aims to minimize the saliency of pruned elements, thereby optimizing the sparsity pattern achieved through permutation. The cost function is articulated as follows:

\begin{equation}
C_{i, j}=\rho - |M \odot \rho|, \text{ where } \rho \subset P_i \cup s_j
\end{equation}

Where $P_i$ represents the $i$-th partition, and $s_j$ is the $j$-th sample (or cluster). The term $\rho$ denotes the saliency scores of the weights, and $M$ is the mask reflecting the target sparsity pattern driven by the permutation. The primary goal of this cost function is to minimize the saliency of the weights that are pruned according to the targeted pruning method when the $j$-th sample is integrated into the $i$-th partition.

After calculating the cost for all combinations of partitions and samples, the Hungarian algorithm is employed to find the combination that minimizes the total cost. This pre-optimization of the cost during the permutation process makes it possible to specifically target the least important elements during actual pruning operations. By preemptively optimizing the permutations in this manner, the method effectively reduces the impact of removing critical elements, enhancing the overall efficiency and effectiveness of the pruning process.

\section{Experiments}
For our model selection, we opted for ResNet18, ResNet50, DeiT-base, and BERT-base as experimental models, demonstrating the effectiveness of the HiNM technique through enhancements in accuracy and F1 scores. We utilized the ImageNet-1K dataset for the CNN and DeiT-base models, and the SQuAD 1.1 dataset for the BERT model.

In Section~\ref{Efficiency by Gyro-Permutation}, we first evaluate the accuracy improvement by gyro-permutation under different sparsity ratios. We examined two scenarios: \textit{one-shot pruning with fine-tuning} and \textit{gradual pruning}. 
In Section~\ref{Comparing Other Permutation Methods}, we conduct an ablation study to assess the superiority of our permutation techniques over previously established methods.
In Section~\ref{Latency Overhead by Gyro-Permutation}, we benchmark the end-to-end latency of the inference task using the BERT-base model to ascertain the speed improvements facilitated by HiNM pruning. The experiments were conducted on an NVIDIA RTX 4090 GPU, leveraging its Ampere architecture.

\subsection{Accuracy improvement by Gyro-permutation} \label{Efficiency by Gyro-Permutation}

To assess the efficiency of gyro-permutation, we begin by employing one-shot pruning complemented by fine-tuning. To demonstrate the superiority of HiNM, we compare its outcomes with those of the vector-wise sparsity method~\cite{ovw} and the element-wise pruning method.

For estimating the importance of weight (or vector) elements, two approaches can be utilized: the magnitude (L1 norm) technique~\cite{deepcomp} and the second-order information method~\cite{lecun1989optimal, kurtic2022optimal}. We apply the simpler magnitude technique for CNN models due to its straightforwardness. However, for the transformer-based DeiT-base model, we employ the second-order technique, considering the increased complexity  of transformer models.


\subsubsection{One-shot pruning with fine-tuning} \label{One-shot pruning with fine-tuning}

{\bf{Environments.}}
We apply HiNM pruning to all the {\verb|Conv2d|} layers, setting the vector size to 32 for ResNet models sourced from the torchvision library~\cite{marcel2010torchvision}. During the fine-tuning phase, we utilize a cosine learning rate strategy with a value of 0.05 over 60 epochs. 
For the DeiT-base model, we implement the second-order pruning across all {\verb|Linear|} modules within the attention, intermediate, and output layers. We adopt an exponential learning rate scheme set at \(10^{-4}\)  and continue fine-tuning for 60 epochs.

\begin{figure}[htbp]
\centering
\begin{minipage}[t]{0.48\textwidth}
\centering
  \includegraphics[width=6.5cm]{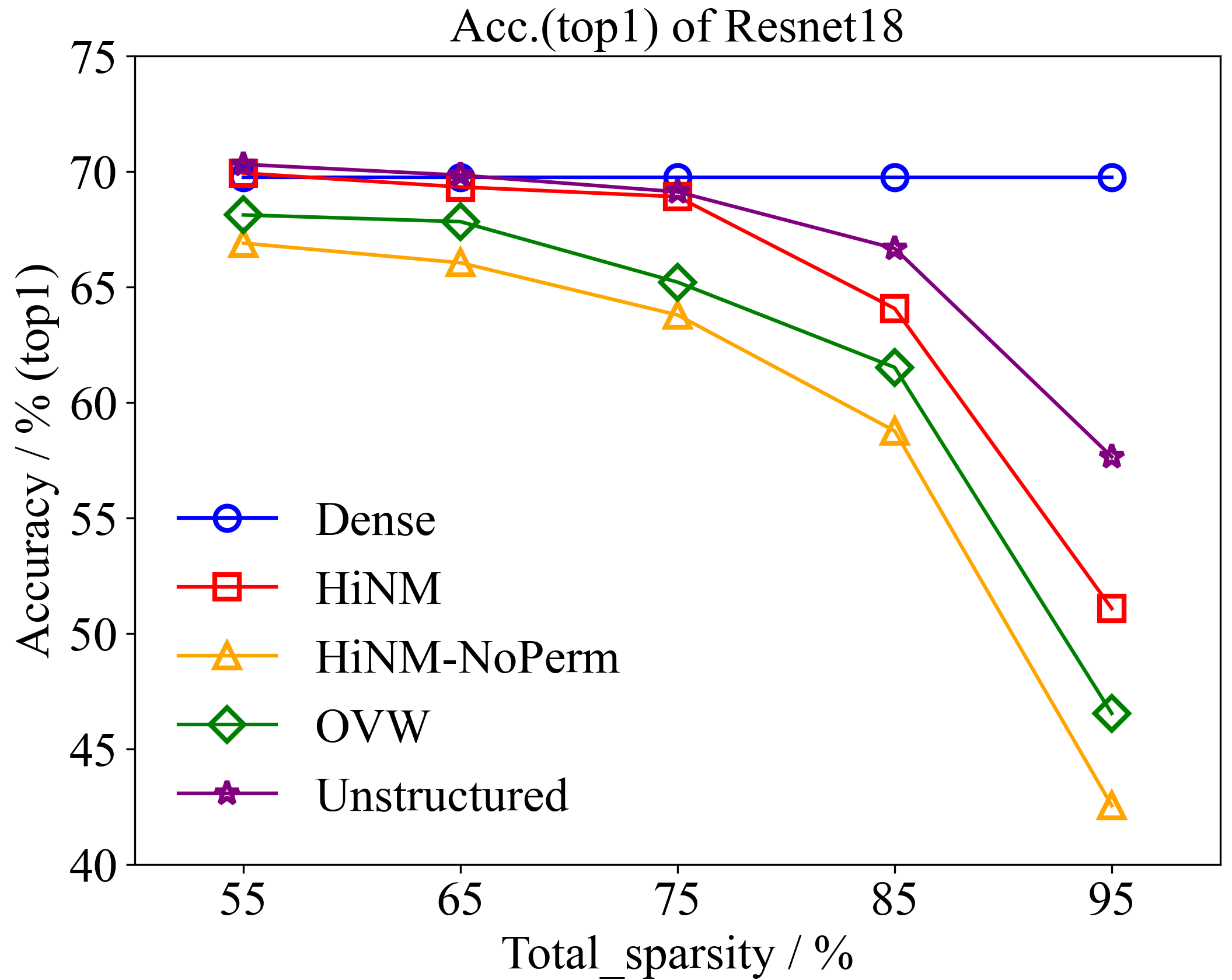}
  \caption{One-shot pruning for ResNet18}
  \label{One-shot pruning for ResNet18}
\end{minipage}
\begin{minipage}[t]{0.48\textwidth}
\centering
  \includegraphics[width=6.5cm]{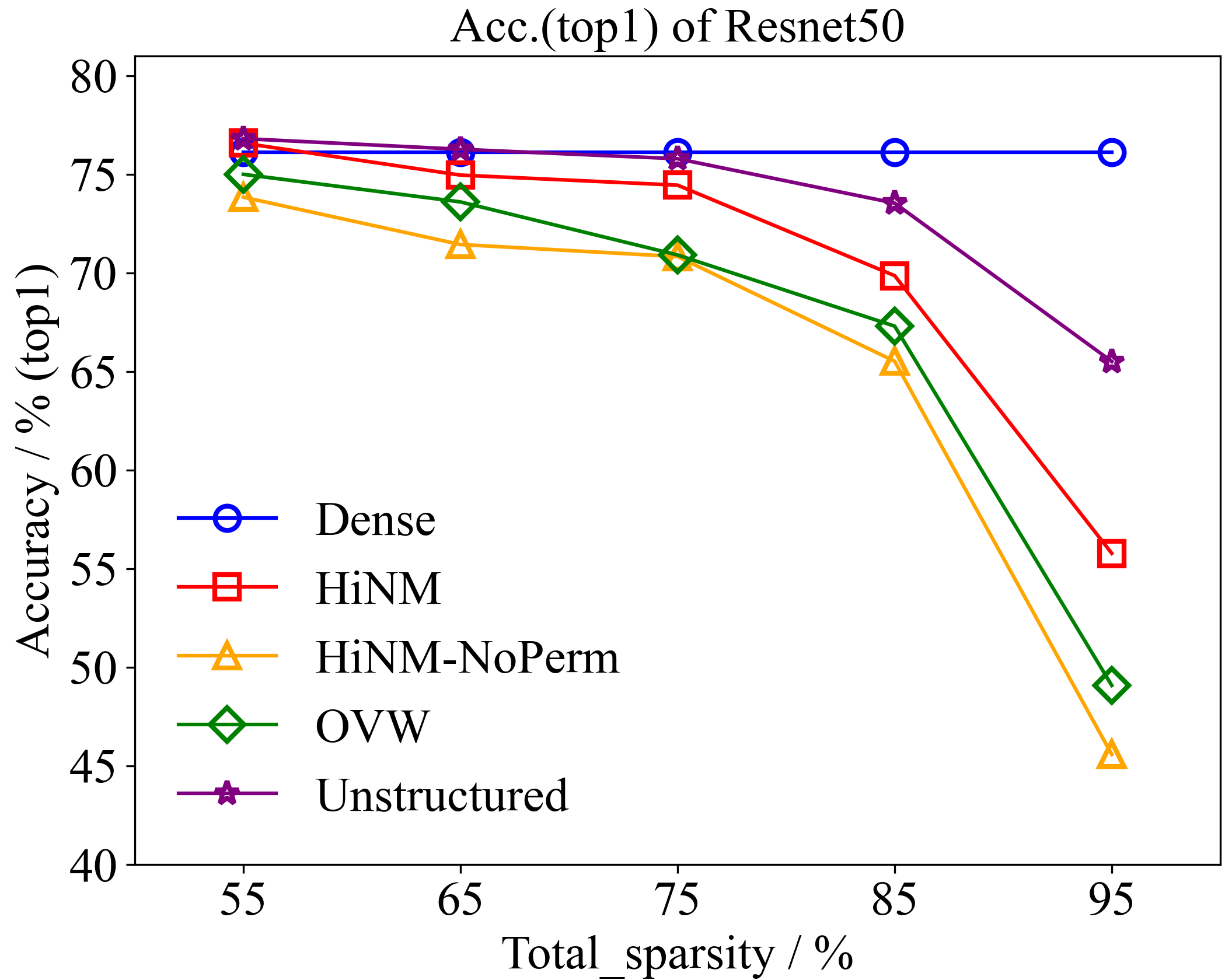}
  \caption{One-shot pruning for ResNet50}
  \label{One-shot pruning for ResNet50}
\end{minipage}
\end{figure}

We evaluate HiNM against four other pruning techniques. As indicated in the legends of Figures~\ref{One-shot pruning for ResNet18} and~\ref{One-shot pruning for ResNet50},
\texttt{Dense} refers to the accuracy of the original, unpruned model;
\texttt{HiNM-NoPerm} describes a variant of our proposed strategy that omits the gyro-permutation component;
\texttt{OVW} represents traditional out-vector-wise sparse pattern pruning~\cite{ovw};
\texttt{Unstructured} corresponds to element-wise pruning.

The results indicate a notable decline in accuracy for HiNM pruning without permutation as the sparsity ratio increases. However, integrating permutation technology significantly enhances accuracy, restoring it to levels comparable to those of unstructured pruning. Particularly at a 75\% sparsity rate, our gyro channel permutation technique improves accuracy by 5.12\% for ResNet18 and 3.62\% for ResNet50.

Notably, even with finer-grained N:M sparsity, HiNM-NoPerm underperforms compared to OVW, which benefits from a channel permutation technique tailored for vector-wise pruning. Nevertheless, the introduction of gyro-permutation in HiNM enables it to surpass OVW in accuracy.

Specifically, at a sparsity rate of 75\%, HiNM achieves an accuracy of 68.91\% on ResNet18 and 74.45\% on ResNet50, exceeding the OVW's accuracy of 65.21\% and 70.91\%, respectively, as shown in Figures~\ref{One-shot pruning for ResNet18} and~\ref{One-shot pruning for ResNet50}. Consequently, HiNM sparsity maintains approximately 99\% of the dense ResNet18 model's accuracy and 98\% of the dense ResNet50 model's accuracy, even with a 75\% reduction in model complexity.

\begin{table}[t]
\centering
\begin{minipage}[t]{0.48\textwidth}
\centering
  \caption{One-shot pruning for DeiT-base}
  \label{One-shot pruning for DeiT-base}
    \small{
    \begin{tabular}{l|lll}
    \hline
    \multirow{2}{*}{Method} & \multicolumn{3}{c}{Sparsity (\%)} \\
                            & 65     & 75    & 85    \\ \hline
    Dense                   & 81.80  & 81.80 & 81.80 \\
    HiNM                    & 81.37  & 81.14 & 75.30 \\
    HiNM-NoPerm             & 77.30  & 76.10 & 63.11 \\
    CAP                     & 81.29  & 81.00 & 74.52 \\ \hline
    \end{tabular}
    }
\end{minipage}
\begin{minipage}[t]{0.48\textwidth}
\centering
  \caption{Gradual pruning for Bert-base: F1 score}
  \label{Gradual pruning for Bert-base: F1 score}
\begin{tabular}{l|ll}
\hline
\multicolumn{1}{l|}{\multirow{2}{*}{Model}} & \multicolumn{2}{c}{Sparsity (\%)} \\
\multicolumn{1}{c|}{}                       & 75          & 87.5         \\ \hline
HiNM                                        & 88.04       & 85.79      \\
VENOM                                       & 87.23       & 84.86      \\ \hline
\end{tabular}
\end{minipage}
\end{table}

Table~\ref{One-shot pruning for DeiT-base} presents the results for the DeiT-base model. Our HiNM technique utilizes the second-order information pruning and is benchmarked against the state-of-the-art (SOTA) element-level weight pruning technique, CAP~\cite{kuznedelev2024cap}, which features weight correlation-aware second-order pruning technology. 
The data indicate that HiNM outperforms all other methods. Specifically, HiNM shows a 5.04\% improvement over HiNM-NoPerm and a 0.14\% improvement over CAP at a 75\% sparsity level. 
These experiments demonstrate that the integration of gyro-permutation with HiNM effectively minimizes unnecessary losses during the pruning process.

\subsubsection{Gradual pruning} \label{Gradual pruning}

{\bf{Environments}}
We conducted gradual pruning on the Bert-base model to compare its accuracy with VENOM~\cite{venom}, which employs the same sparsity pattern. VENOM adjusts its saliency scores using a pair-wise approach within a second-order pruning technique~\cite{obert} and modifies both the vector sparsity ratio and N:M sparsity ratio with each gradual pruning step. In contrast, to demonstrate the efficiency of gyro-permutation, we based our permutations on saliency scores using a second-order pruning technique and aligned our pruning steps with the HiNM sparsity pattern. Initially, we applied only column-wise vector pruning during the early stages of gradual pruning. Once the target vector sparsity ratio is achieved, we then proceeded with N:M pruning. According to Table~\ref{Gradual pruning for Bert-base: F1 score}, the F1 score for the HiNM sparsity with gyro-permutation showed improvements of 0.81\% and 0.93\% over VENOM’s performance in the Bert-base model.

\subsection{Ablation study on various permutation methods} \label{Comparing Other Permutation Methods}
Our gyro-permutation technique features innovative approaches in both output channel permutation (OCP) and tile-wise input channel permutation (ICP), distinguishing it from previous single-level permutation techniques that were limited to either output or input channels and often faced issues with the probability of local minima. To assess the efficiency of individual permutation algorithms within our gyro-permutation framework, we conducted an ablation study. This study involved substituting the OCP and ICP with prior permutation techniques, leading to the creation of two variants: HiNM-V1, which modifies the OCP, and HiNM-V2, which alters the ICP.

{\bf{HiNM-V1}}
Unlike our OCP, where the number of samples varies across iterations, the OVW does not involve a sampling phase and instead utilizes a balanced K-means clustering algorithm to partition all output channels into each group uniformly. Consequently, as illustrated in Table~\ref{Ablation study}, our HiNM achieves an accuracy improvement of 4.53\% and 0.49\% over HiNM-V1 for ResNet18 and ResNet50, respectively. This performance enhancement can be attributed to the factors discussed in Section~\ref{sec:Gyro}. Specifically, clustering more channels in a single iteration during the permutation process complicates the identification of the optimal order. Moreover, the balanced K-means clustering algorithm does not explicitly identify elements to be pruned, which can lead to significant accuracy drop if used alone.





{\bf{HiNM-V2}}
The input channel permutation technique from NVIDIA-Apex~\cite{nvidia_channelperm} rearranges input channels according to their target N:M sparsity pattern. To benchmark against this technique, we altered the granularity of the permutation unit from input channels to column vectors. As shown in Table~\ref{Ablation study}, the accuracy of HiNM exceeds that of HiNM-V2 by 2.5\% and 0.87\%, respectively. This suggests that a small sample size can be prone to falling into local optima, even when using strategies from ~\cite{nvidia_channelperm} designed to avoid such pitfalls.




\begin{table}[htp]
\centering
  \caption{Ablation study}
  \label{Ablation study}
\begin{tabular}{c|ll|c|ll}
\hline
\multicolumn{1}{l|}{Model} & Method  & Top1 acc. & \multicolumn{1}{l|}{Model} & Method  & Top1 acc. \\ \hline
\multirow{3}{*}{Resnet18}  & HiNM    & 68.91   & \multirow{3}{*}{Resnet50}  & HiNM    & 74.45   \\
                           & HiNM-V1 & 64.38   &                            & HiNM-V1 & 73.96   \\
                           & HiNM-V2 & 66.41   &                            & HiNM-V2 & 73.58   \\ \hline
\end{tabular}
\end{table}

Compared to HiNM, HiNM-V1 experienced a decrease in accuracy of 4.53\% on the ResNet18 model and no change on the ResNet50 model. Similarly, there exists the accuracy drop of 2.5\% on the ResNet18 model and 0.87\% on the ResNet50 model when applying HiNM-V1. 
It indicates that regardless of the direction of channel permutation, our algorithm exhibits significant preference.

\subsection{Latency overhead by Gyro-permutation} \label{Latency Overhead by Gyro-Permutation}

\begin{figure}[htb]
  \centering
  \includegraphics[width=.8\textwidth]{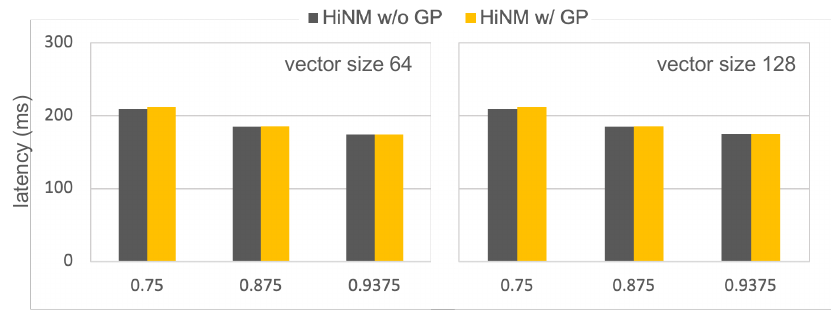}
  \caption{Latency overhead of Gyro-Permutation of Bert-base}
  \label{latency}
\end{figure}

We adapted the GPU kernel previously utilized by VENOM~\cite{venom} to accommodate HiNM sparsity with gyro-permutation. In contrast to VENOM, which employed padding techniques to mitigate bank conflict during the storage of partial sums in shared memory, we implemented NVIDIA's swizzle operator to address this conflict. Additionally, for maintaining inter-layer consistency, the output channel permutation of gyro-permutation systematically prearranges all inter-layer channels offline. Meanwhile, the input channel permutation dynamically adjusts vector indices at runtime. To assess the runtime overhead of gyro-permutation across various sparsity ratios, we integrated HiNM sparsity into the Bert-base model on a RTX 3090 GPU and recorded the inference times. As illustrated in Figure~\ref{latency}, gyro-permutation introduces no detectable runtime overhead across all tested sparsity ratios and vector sizes.

\section{Conclusions}
In this study, we introduce the novel gyro-permutation technique tailored for HiNM sparsity, specifically reorders both output channels and input channels. This method not only maintains layer-to-layer consistency but also proves the feasibility of synchronizing the ordering of both output and input channels by observing the data movement across column vectors during GPU computations. By methodically sequencing the permutations—initiating with output channels and progressing to tile-wise input channels—we significantly improve the efficiency of both column-wise vector pruning and N:M pruning within a hierarchical pruning framework.

Furthermore, our permutation algorithm, which encompasses sampling, clustering, and assignment phases, brings unique advantages that collectively boost the effectiveness of the entire system.

\section{Limitations}
Due to the diverse pruning policies available within Hierarchical N:M sparsity, the sequence of permutations can vary based on the chosen pruning policy. We adhere to the conventional approach, initiating with column-wise vector pruning before proceeding to N:M pruning. However, we anticipate that future research will demonstrate enhanced performance when pruning policies and permutations are strategically integrated.

\bibliographystyle{IEEEtran}
\bibliography{ref}

\begin{thebibliography}{10}
\providecommand{\url}[1]{#1}
\csname url@samestyle\endcsname
\providecommand{\newblock}{\relax}
\providecommand{\bibinfo}[2]{#2}
\providecommand{\BIBentrySTDinterwordspacing}{\spaceskip=0pt\relax}
\providecommand{\BIBentryALTinterwordstretchfactor}{4}
\providecommand{\BIBentryALTinterwordspacing}{\spaceskip=\fontdimen2\font plus
\BIBentryALTinterwordstretchfactor\fontdimen3\font minus \fontdimen4\font\relax}
\providecommand{\BIBforeignlanguage}[2]{{%
\expandafter\ifx\csname l@#1\endcsname\relax
\typeout{** WARNING: IEEEtran.bst: No hyphenation pattern has been}%
\typeout{** loaded for the language `#1'. Using the pattern for}%
\typeout{** the default language instead.}%
\else
\language=\csname l@#1\endcsname
\fi
#2}}
\providecommand{\BIBdecl}{\relax}
\BIBdecl

\bibitem{scaling_law}
J.~Kaplan, S.~McCandlish, T.~Henighan, T.~B. Brown, B.~Chess, R.~Child, S.~Gray, A.~Radford, J.~Wu, and D.~Amodei, ``Scaling laws for neural language models,'' \emph{arXiv preprint arXiv:2001.08361}, 2020.

\bibitem{han_unstructured}
S.~Han, J.~Pool, J.~Tran, and W.~Dally, ``Learning both weights and connections for efficient neural network,'' \emph{Advances in neural information processing systems}, vol.~28, 2015.

\bibitem{channel_pruning}
Y.~He, X.~Zhang, and J.~Sun, ``Channel pruning for accelerating very deep neural networks,'' in \emph{Proceedings of the IEEE international conference on computer vision}, 2017, pp. 1389--1397.

\bibitem{ovw}
Y.~Tan, K.~Han, K.~Zhao, X.~Yu, Z.~Du, Y.~Chen, Y.~Wang, and J.~Yao, ``Accelerating sparse convolution with column vector-wise sparsity,'' \emph{Advances in Neural Information Processing Systems}, vol.~35, pp. 30\,307--30\,317, 2022.

\bibitem{filter_pruning}
H.~Li, A.~Kadav, I.~Durdanovic, H.~Samet, and H.~P. Graf, ``Pruning filters for efficient convnets,'' in \emph{International Conference on Learning Representations}, 2016.

\bibitem{STC}
A.~Mishra, J.~A. Latorre, J.~Pool, D.~Stosic, D.~Stosic, G.~Venkatesh, C.~Yu, and P.~Micikevicius, ``Accelerating sparse deep neural networks,'' \emph{arXiv preprint arXiv:2104.08378}, 2021.

\bibitem{shfl-bw}
G.~Huang, H.~Li, M.~Qin, F.~Sun, Y.~Ding, and Y.~Xie, ``Shfl-bw: accelerating deep neural network inference with tensor-core aware weight pruning,'' in \emph{Proceedings of the 59th ACM/IEEE Design Automation Conference}, 2022, pp. 1153--1158.

\bibitem{nvidia_channelperm}
J.~Pool and C.~Yu, ``Channel permutations for n: m sparsity,'' \emph{Advances in neural information processing systems}, vol.~34, pp. 13\,316--13\,327, 2021.

\bibitem{deepcomp}
S.~Han, H.~Mao, and W.~J. Dally, ``Deep compression: Compressing deep neural networks with pruning, trained quantization and huffman coding,'' in \emph{International Conference on Learning Representations}, 2016.

\bibitem{movement}
V.~Sanh, T.~Wolf, and A.~Rush, ``Movement pruning: Adaptive sparsity by fine-tuning,'' \emph{Advances in neural information processing systems}, vol.~33, pp. 20\,378--20\,389, 2020.

\bibitem{mfac}
E.~Frantar, E.~Kurtic, and D.~Alistarh, ``M-fac: Efficient matrix-free approximations of second-order information,'' \emph{Advances in Neural Information Processing Systems}, vol.~34, pp. 14\,873--14\,886, 2021.

\bibitem{obert}
E.~Kurtic, D.~Campos, T.~Nguyen, E.~Frantar, M.~Kurtz, B.~Fineran, M.~Goin, and D.~Alistarh, ``The optimal bert surgeon: Scalable and accurate second-order pruning for large language models,'' in \emph{Proceedings of the 2022 Conference on Empirical Methods in Natural Language Processing}, 2022, pp. 4163--4181.

\bibitem{oneshot}
N.~Lee, T.~Ajanthan, and P.~Torr, ``{SNIP}: {SINGLE}-{SHOT} {NETWORK} {PRUNING} {BASED} {ON} {CONNECTION} {SENSITIVITY},'' in \emph{International Conference on Learning Representations}, 2019.

\bibitem{balanced_vw}
C.~Park, M.~Park, H.~J. Oh, M.~Kim, M.~K. Yoon, S.~Kim, and W.~W. Ro, ``Balanced column-wise block pruning for maximizing gpu parallelism,'' in \emph{Proceedings of the AAAI Conference on Artificial Intelligence}, vol.~37, no.~8, 2023, pp. 9398--9407.

\bibitem{learning_nm}
A.~Zhou, Y.~Ma, J.~Zhu, J.~Liu, Z.~Zhang, K.~Yuan, W.~Sun, and H.~Li, ``Learning n: M fine-grained structured sparse neural networks from scratch,'' in \emph{International Conference on Learning Representations}, 2021.

\bibitem{transposable_nm}
I.~Hubara, B.~Chmiel, M.~Island, R.~Banner, J.~Naor, and D.~Soudry, ``Accelerated sparse neural training: A provable and efficient method to find n: m transposable masks,'' \emph{Advances in neural information processing systems}, vol.~34, pp. 21\,099--21\,111, 2021.

\bibitem{domino}
W.~Sun, A.~Zhou, S.~Stuijk, R.~Wijnhoven, A.~O. Nelson, H.~Corporaal \emph{et~al.}, ``Dominosearch: Find layer-wise fine-grained n: M sparse schemes from dense neural networks,'' \emph{Advances in neural information processing systems}, vol.~34, pp. 20\,721--20\,732, 2021.

\bibitem{dissecting}
W.~Sun, A.~Li, T.~Geng, S.~Stuijk, and H.~Corporaal, ``Dissecting tensor cores via microbenchmarks: Latency, throughput and numeric behaviors,'' \emph{IEEE Transactions on Parallel and Distributed Systems}, vol.~34, no.~1, pp. 246--261, 2022.

\bibitem{highlight}
Y.~N. Wu, P.-A. Tsai, S.~Muralidharan, A.~Parashar, V.~Sze, and J.~Emer, ``Highlight: Efficient and flexible dnn acceleration with hierarchical structured sparsity,'' in \emph{Proceedings of the 56th Annual IEEE/ACM International Symposium on Microarchitecture}, 2023, pp. 1106--1120.

\bibitem{tstc}
J.~Liu, G.~Dai, H.~Xia, L.~Guo, X.~Shi, J.~Xu, H.~Yang, and Y.~Wang, ``Tstc: Two-level sparsity tensor core enabling both algorithm flexibility and hardware efficiency,'' in \emph{2023 IEEE/ACM International Conference on Computer Aided Design (ICCAD)}.\hskip 1em plus 0.5em minus 0.4em\relax IEEE, 2023, pp. 1--9.

\bibitem{venom}
R.~L. Castro, A.~Ivanov, D.~Andrade, T.~Ben-Nun, B.~B. Fraguela, and T.~Hoefler, ``Venom: A vectorized n: M format for unleashing the power of sparse tensor cores,'' in \emph{Proceedings of the International Conference for High Performance Computing, Networking, Storage and Analysis}, 2023, pp. 1--14.

\bibitem{tetris}
Y.~Ji, L.~Liang, L.~Deng, Y.~Zhang, Y.~Zhang, and Y.~Xie, ``Tetris: Tile-matching the tremendous irregular sparsity,'' \emph{Advances in neural information processing systems}, vol.~31, 2018.

\bibitem{lecun1989optimal}
Y.~LeCun, J.~Denker, and S.~Solla, ``Optimal brain damage,'' \emph{Advances in neural information processing systems}, vol.~2, 1989.

\bibitem{kurtic2022optimal}
E.~Kurtic, D.~Campos, T.~Nguyen, E.~Frantar, M.~Kurtz, B.~Fineran, M.~Goin, and D.~Alistarh, ``The optimal bert surgeon: Scalable and accurate second-order pruning for large language models,'' in \emph{Proceedings of the 2022 Conference on Empirical Methods in Natural Language Processing}, 2022, pp. 4163--4181.

\bibitem{marcel2010torchvision}
S.~Marcel and Y.~Rodriguez, ``Torchvision the machine-vision package of torch,'' in \emph{Proceedings of the 18th ACM international conference on Multimedia}, 2010, pp. 1485--1488.

\bibitem{kuznedelev2024cap}
D.~Kuznedelev, E.~Kurti{\'c}, E.~Frantar, and D.~Alistarh, ``Cap: Correlation-aware pruning for highly-accurate sparse vision models,'' \emph{Advances in Neural Information Processing Systems}, vol.~36, 2024.

\end{thebibliography}


\end{document}